\documentclass[lettersize,journal]{IEEEtran}
\usepackage{amsmath,amsfonts}
\usepackage{algorithmic}
\usepackage{algorithm}
\usepackage{array}
\usepackage{booktabs}
\usepackage{multirow}
\usepackage[caption=false,font=normalsize,labelfont=sf,textfont=sf]{subfig}
\usepackage{textcomp}
\usepackage{stfloats}
\usepackage{url}
\usepackage{verbatim}
\usepackage{graphicx}
\usepackage{cite}
\def\ie{{\em i.e.}}

\usepackage{xcolor}
\usepackage{color}
\usepackage{bm}
\usepackage{bbm}
\usepackage{dsfont}

\begin{document}

\title{General vs. Long-Tailed Age Estimation: An Approach to Kill Two Birds with One Stone}

\author{Zenghao Bao, Zichang Tan,~\IEEEmembership{Member,~IEEE,} Jun Li, Jun Wan,~\IEEEmembership{Senior Member,~IEEE,} \\ Xibo Ma, and Zhen Lei,~\IEEEmembership{Senior Member,~IEEE}

\thanks{This work was supported by the National Key Research and Development Plan under Grant 2021YFE0205700, the External cooperation key project of Chinese Academy Sciences  173211KYSB20200002, the Chinese National Natural Science Foundation Projects 62276254, 62176256, 62106264 and 82090051, the Science and Technology Development Fund of Macau Project 0123/2022/A3, 0070/2020/AMJ, Guangdong Provincial Key R\&D Programme: 2019B010148001, CCF-Zhipu AI Large Model OF 202219, Open Research Projects of Zhejiang Lab No. 2021KH0AB01, and the InnoHK program. \emph{(Jun Wan is Corresponding authors.)}}
\thanks{Z. Bao, J. Li, and X. Ma are with the State Key Laboratory of Multimodal Artificial Intelligence Systems (MAIS), Institute of Automation, Chinese Academy of Sciences (CASIA), Beijing 100190, China, also with the School of Artificial Intelligence, University of Chinese Academy of Sciences (UCAS), Beijing 100049, China. E-mail: \{baozenghao2020, lijun2021\}@ia.ac.cn, xibo.ma@nlpr.ia.ac.cn.}
\thanks{J. Wan is with the State Key Laboratory of Multimodal Artificial Intelligence Systems (MAIS), Institute of Automation, Chinese Academy of Sciences (CASIA), Beijing 100190, China, also with the School of Artificial Intelligence, University of Chinese Academy of Sciences (UCAS), Beijing 100049, China, and also with Macau University of Science and Technology, Macau 999078, China. E-mail: jun.wan@ia.ac.cn}
\thanks{Z. Tan is with the Institute of Deep Learning, Baidu Research, Beijing, 100085, China. E-mail: tanzichang@baidu.com.}
\thanks{Z. Lei is with the State Key Laboratory of Multimodal Artificial Intelligence Systems (MAIS), Institute of Automation, Chinese Academy of Sciences (CASIA), Beijing 100190, China, also with the School of Artificial Intelligence, University of Chinese Academy of Sciences (UCAS), Beijing 100049, China, and also with the Centre for Artificial Intelligence and Robotics, Hong Kong Institute of Science \& Innovation, Chinese Academy of Sciences, Hong Kong. Email: zlei@nlpr.ia.ac.cn.}}

\markboth{IEEE TRANSACTIONS ON IMAGE PROCESSING, VOL. XX, NO. XX, 2022}%
{Shell \MakeLowercase{\textit{et al.}}: A Sample Article Using IEEEtran.cls for IEEE Journals}


\maketitle

\begin{abstract}
Facial age estimation has received a lot of attention for its diverse application scenarios. 
Most existing studies treat each sample equally and aim to reduce the average estimation error for the entire dataset, which can be summarized as \textit{\textbf{General Age Estimation}}. 
However, due to the long-tailed distribution prevalent in the dataset, treating all samples equally will inevitably bias the model toward the head classes (usually the adult with a majority of samples).
Driven by this, some works suggest that each class should be treated equally to improve performance in tail classes (with a minority of samples), which can be summarized as \textit{\textbf{Long-tailed Age Estimation}}.
However, Long-tailed Age Estimation usually faces a performance trade-off, \ie, achieving improvement in tail classes by sacrificing the head classes.
In this paper, our goal is to design a unified framework to perform well on both tasks, killing two birds with one stone.
To this end, we propose a simple, effective, and flexible training paradigm named GLAE, which is two-fold. 
First, we propose Feature Rearrangement (FR) and Pixel-level Auxiliary learning (PA) for better feature utilization to improve the overall age estimation performance. 
Second, we propose Adaptive Routing (AR) for selecting the appropriate classifier to improve performance in the tail classes while maintaining the head classes. 
Moreover, we introduce a new metric, named Class-wise Mean Absolute Error (CMAE), to equally evaluate the performance of all classes. 
Our GLAE provides a surprising improvement on Morph II, reaching the lowest MAE and CMAE of 1.14 and 1.27 years, respectively. Compared to the previous best method, MAE dropped by up to 34\%, which is an unprecedented improvement, and for the first time, MAE is close to 1 year old. Extensive experiments on other age benchmark datasets, including CACD, MIVIA, and Chalearn LAP 2015, also indicate that GLAE outperforms the state-of-the-art approaches significantly. 
\end{abstract}




\begin{figure}[!t]
\centering
\includegraphics[width=1.0\linewidth]{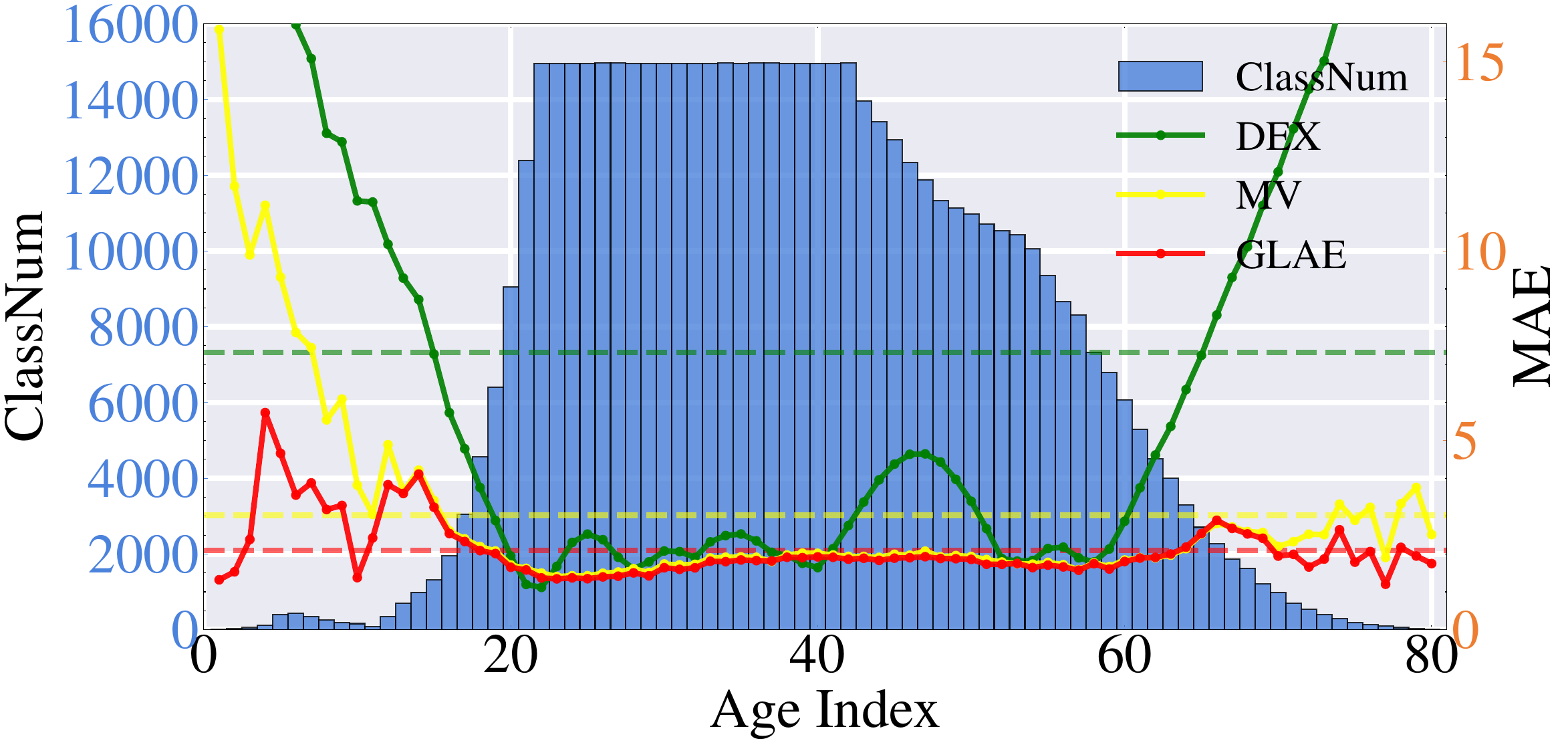}
\caption{\textbf{Comparison with the previous state-of-the-art methods on the MIVIA dataset}. The blue bars indicate the number of samples per age category, the different solid lines indicate the MAE of the different methods in each age category, and the dashed lines indicate the CMAE results of DEX~\cite{rothe2018deep} (green color), MV~\cite{pan2018mean} (yellow color) and our proposed GLAE (red color). Our GLAE outperforms the other methods in both the General Age Estimation (MAE) and Long-tailed Age Estimation (CMAE).
}
\label{fig_intro}
\end{figure}

\section{Introduction}
\IEEEPARstart{F}{acial} age estimation~\cite{duan2020egroupnet,deng2021pml,bao2021lae} refers to identifying a person's age (accumulated years after birth) from his/her face image. This task has received a lot of attention due to its wide range of applications, such as video surveillance, image retrieval and human-computer interaction.
In the past years, facial age estimation has grown by leaps and bounds, achieving considerable improvement with deep learning methods~\cite{yang2018ssr,gao2018age,li2020deep}. However, facial age estimation remains challenging due to the various complex factors in facial aging (e.g., genes, work and living environment)~\cite{tan_pami_2018,gao2018age,tan2019deeply}.

Previous works~\cite{pan2020self,liu2020similarity,duan2020age} address the problem of age estimation mainly from the following aspects, including designing the network architecture~\cite{yang2018ssr,duan2020egroupnet,tan2019deeply},
modifying the loss function~\cite{li2018deep,gao2018age}, and improving the training strategy~\cite{li2020deep,bao2021lae}.
However, these methods ignore the impact of data distribution and treat each sample equally in the training process.
These traditional methods are summarized by us as \textit{\textbf{General Age Estimation }}.
Specifically, the number of people of different ages is imbalanced in real-world scenarios.
Not to mention that in specific application scenarios, the number of people of a certain age will be extremely large/small.
For example, the age detection system of the amusement park will collect a large number of video information of children and young people, and the elderly will rarely appear.
This imbalanced distribution is also reflected in the artificially collected age datasets.
As shown in Fig.~\ref{fig_intro}, the data distribution of the MIVIA dataset is extremely imbalanced, with the category with the largest number of training samples having 15,000 images, while the minimum is only 17.
Previous works~\cite{wang2017learning,bao2021lae} have explored this imbalanced distribution and summarized it as the long-tailed distribution.
 When training on a dataset with a long-tailed distribution, the model will be dominated by the head classes. In other words, the model can achieve low errors in head classes but perform poorly in tail classes, which can be reflected in Fig.~\ref{fig_intro}.
This phenomenon hinders the practical application of age estimation technology. To address this, \textit{\textbf{Long-tailed Age Estimation}}~\cite{bao2021lae} is proposed to train a balanced estimator, which aims to narrow the performance gap between head and tail classes. However, Long-tailed Age Estimation usually faces a performance trade-off, \ie, the improvement of the tail classes inevitably brings the decline of the head classes. 
Both General Age Estimation and Long-tailed Age Estimation are crucial for real-world applications of age estimation, where General Age Estimation helps achieve a higher ceiling when facing balanced data distribution, while Long-tailed Age Estimation will be more stable when dealing with specific, extreme imbalanced distribution scenarios. However, previous studies only considered General Age Estimation or Long-tailed Age Estimation, not both.
\textit{\textbf{
In this paper, we propose a unified framework to perform well on both tasks}} (as shown in Fig~\ref{fig_intro}), \textit{\textbf{killing two birds with one stone.}}

The key to improving General Age Estimation is to learn high-quality representations.
Previous age estimators~\cite{gao2018age,pan2018mean,bao2021lae} are usually constructed on Convolutional Neural Networks (CNN),
where Global Averaging Pooling (GAP) is employed to pool the feature map to a single vector (each channel of feature map is reduced to an element).
Undoubtedly, such reduction will result in the information loss, especially for the local and detailed features, which are crucial for facial age estimation~\cite{tan2019deeply}.
To address the above issues, we first utilize a Feature Rearrangement (FR) to fully make use of the visual information carried by feature maps. To be specific, the pixels of the same location will be collected and placed together in the re-arranged feature map, which is followed by a delicate convolutional layer to fully capture the correlation of pixels of the same location. Moreover, in traditional CNN architectures, the feature is often followed by a linear classifier to obtain scores corresponding to each category. We utilize our well-designed convolution layer to directly obtain the score of each pixel on the feature map in each category, as an alternative to a linear classifier. In order to further make better use of the information of overall features and pixel-level features, a Pixel-level Auxiliary learning (PA) is further proposed to capture meticulous features. The proposed PA consists of two branches to calculate the loss independently. One basically follows the traditional way in image recognition but uses an MLP projection to replace GAP to achieve global and robust learning for all pixels, reducing the information loss caused by GAP. The other is to conduct the learning in a more detailed way. The feature map will be split pixel by pixel, and pixel-level prediction scores will be utilized to aid training. In this way, the local and meticulous features will be captured to cooperate with the global feature for high-quality representation.


When training the age estimator in the long-tailed scenario, its classifier is usually biased towards the head classes, where the model achieves better performance on head classes than that of tail classes.
One way to alleviate the bias is to retrain the classifier with class-balanced sampling (\ie, each class has the same probability to be sampled)~\cite{kang2019decoupling}. 
As clarified in previous works~\cite{bao2021lae}, such a training paradigm could improve the performance of tail classes, but it also sacrifices the performance of head classes. 
Obviously, it is not suitable to the practical use of age estimation, where most people involved are adults (\ie head classes). 
As we have mentioned above, the vanilla classifier performs well on head classes, which is complementary to the balanced retrained classifier (balanced classifier).
So can we combine them to achieve good performance in both head and tail classes?
Grounded on this question, we propose an Adaptive Routing (AR) to adaptively select the appropriate classifier for each input image between vanilla and balanced classifiers. 
To be specific, the two classifiers are not activated concurrently for an input image, and only the appropriate classifier is chosen for each input image. Empirically, we want to use the classifier that produces less uncertainty in the output after making small disturbances to the input image. So, our selection criteria is the KL divergence between the predicted distributions of the raw image and the flip copy. The classifier with the smaller KL divergence will be chosen for prediction. Our experiments also demonstrate that our selection achieves better performance.

In the end, we integrate the above modules and propose a training paradigm to achieve good performance for both General and Long-tailed Age Estimation (namely GLAE for short). 
Moreover, there are complete protocols and evaluation criteria for General Age Estimation,
but the protocol and criteria are lacking for Long-tailed Age Estimation because of the few works on Long-tailed Age Estimation.
Due to the extremely imbalanced distribution in age datasets, it is hard to partition a balanced test set that contains enough images for each class.
Therefore, we propose a new metric named Class-wise Mean Absolute Error (CMAE) for long-tailed age estimation, which evaluates the balanced performance over all classes. The main contributions of this paper can be summarized as follows:

\begin{itemize}
\item 
We propose a simple, effective, and flexible training paradigm named GLAE for age estimation. 
The proposed GLAE could achieve good performance on all samples and all classes simultaneously, which does both well for general and long-tailed age estimation.
\item Three novel components, namely Feature Rearrangement (FR), Pixel-level Auxiliary learning (PA), and Adaptive Routing (AR), are proposed to improve the performance of facial age estimation.
\item We propose a new evaluation criterion named CMAE and several protocols to evaluate the performance for long-tailed facial age estimation.
\item Extensive experiments on four popular benchmarks, including Morph II, CACD, MIVIA, and Chalearn LAP 2015, indicate the state-of-the-art performance of GLAE.
\end{itemize}

\section{Related Work}

\subsection{Facial Age Estimation}
Facial age estimation is a very challenging task due to the various complex factors in facial aging~\cite{tan_pami_2018}. In recent years, facial age estimation has been dominated by deep learning based methods~\cite{wan2018auxiliary,zhang2019recurrent,zhang2019fine,chen2019age,li2021learning,xie2019chronological,xie2020deep,liu2020facial,duan2020egroupnet,zhao2020distilling,pan2020self,liu2020similarity,duan2020age} and the performance has been greatly improved.
Most of those works are proposed for general age estimation, and they can be roughly categorized into regression-based~\cite{carletti2019age}, ranking-based~\cite{niu2016ordinal,liu2017ordinal}, classification-based~\cite{liu2017label,rothe2018deep}, and label distribution-based~\cite{gao2017deep,gao2018age,gao2020learning,deng2020learning,sun2021deep} methods. Regression-based methods~\cite{carletti2019age,yi2014age} predict the age by using a regressor, which treats age as a continuous value. 
However, the regressor~\cite{yi2014age,carletti2019age} hardly learns discriminative aging features. Classification-based methods~\cite{liu2017label,rothe2018deep} treat different ages as independent classes and formulate the age estimation as a multi-class classification problem.
However, the classification-based methods ignore the relative correlation among adjacent ages and also struggle with performance. The ranking-based methods~\cite{niu2016ordinal,liu2017ordinal} introduce the ordinal correlation for age estimation. For example, Chen et al.~\cite{chen2017using} use a series of binary classifiers to obtain ordinal information by judging whether the face is older than a certain age.
However, ranking-based approaches are limited to scalar outputs. 
Recently, label distribution-based methods~\cite{gao2017deep,gao2018age,gao2020learning,deng2020learning,sun2021deep} have attracted much attention. Generally speaking, the age label is usually represented as a Gaussian distribution rather than a standalone label. 

All the aforementioned methods are proposed for general age estimation
with the goal of reducing the average error over the whole dataset.
However, they usually perform poorly on children and the elderly due to ignoring the long-tailed distribution in age datasets. How to effectively learn the model under the long-tailed imbalanced distribution is an important research topic for age estimation.
However, there are only a few works that have noticed this~\cite{bao2021lae,deng2021pml}.
For example, Bao et al.~\cite{bao2021lae} decouple the training process into representation learning and classifier learning, where the latter aims to learn a balanced classifier.
It improves the performance of tail classes (\ie children and the elderly) but also experiences a performance drop in head classes (\ie adults).
Both general and long-tailed age estimation are of great importance for practical uses.
However, all previous works consider them separately. In this paper, we propose a unified framework named GLAE to consider them concurrently and unified, which aims to improve the performance of both head and tail classes.

\subsection{Long-tailed Visual Recognition}
Long-tailed visual recognition~\cite{kang2019decoupling,li2022nested,zhou2020bbn,ren2020balanced,wang2021seesaw,wang2020long,liu2019large,tan2023ncl++} has aroused great interest in recent years. Many methods have been proposed to address the long-tailed learning problem, and they can be roughly divided into five categories, including
re-sampling~\cite{zhang2021bag},
re-weighting~\cite{zhang2021bag,wang2021seesaw}, logit adjustment~\cite{li2021long,menon2020long,zhao2021adaptive}, multi-stage training~\cite{kang2019decoupling,li2021self} and multi-expert learning~\cite{li2022nested,wang2020long}. 
Both data re-sampling~\cite{zhang2021bag} and loss re-weighting~\cite{zhang2021bag,wang2021seesaw} are 
commonly used tools in long-tailed learning. The goal of these methods is to increase the contributions of tail classes while suppressing those of head classes, and therefore, the accuracy of tail classes can be greatly improved. For logit adjustment~\cite{li2021long,menon2020long}, it aims to subtract a positive term from the logit. Generally speaking, the term is related to the frequency of each class, and the overall goal is to emphasize more the tail class and less the head class. For multi-stage learning, one classic methods~\cite{kang2019decoupling} is to decouple the learning procedure into representation learning and classifier learning.
The former is to increase the model's ability to extract discriminative features, and the latter is to re-adjust the classifier to favor the tail class.
However, the classifier learning stage usually sacrifices the performance of the head class. Recently, multi-expert methods~\cite{li2022nested,wang2020long} have attracted much attention and achieved promising performance in long-tailed datasets. 
For example, Li et al.~\cite{li2022nested} propose a Nested Collaborative Learning (NCL), which learns multiple experts concurrently and collaboratively to improve the accuracy.
However, the multi-expert method usually has high complexity and needs a large number of calculations. 

Different from these works, we propose an adaptive routing with two classifiers, with one dealing with the head class and the other for the tail class.
Based on the proposed routing strategy, the appropriate classifier will be selected for each input image to achieve reliable recognition. Most previous methods improve the accuracy of the tail class by sacrificing the accuracy of the head class, while our method improves the accuracy of the tail class while minimizing the impact on the head class.

\begin{figure*}[!t]
\centering
\includegraphics[width=7in]{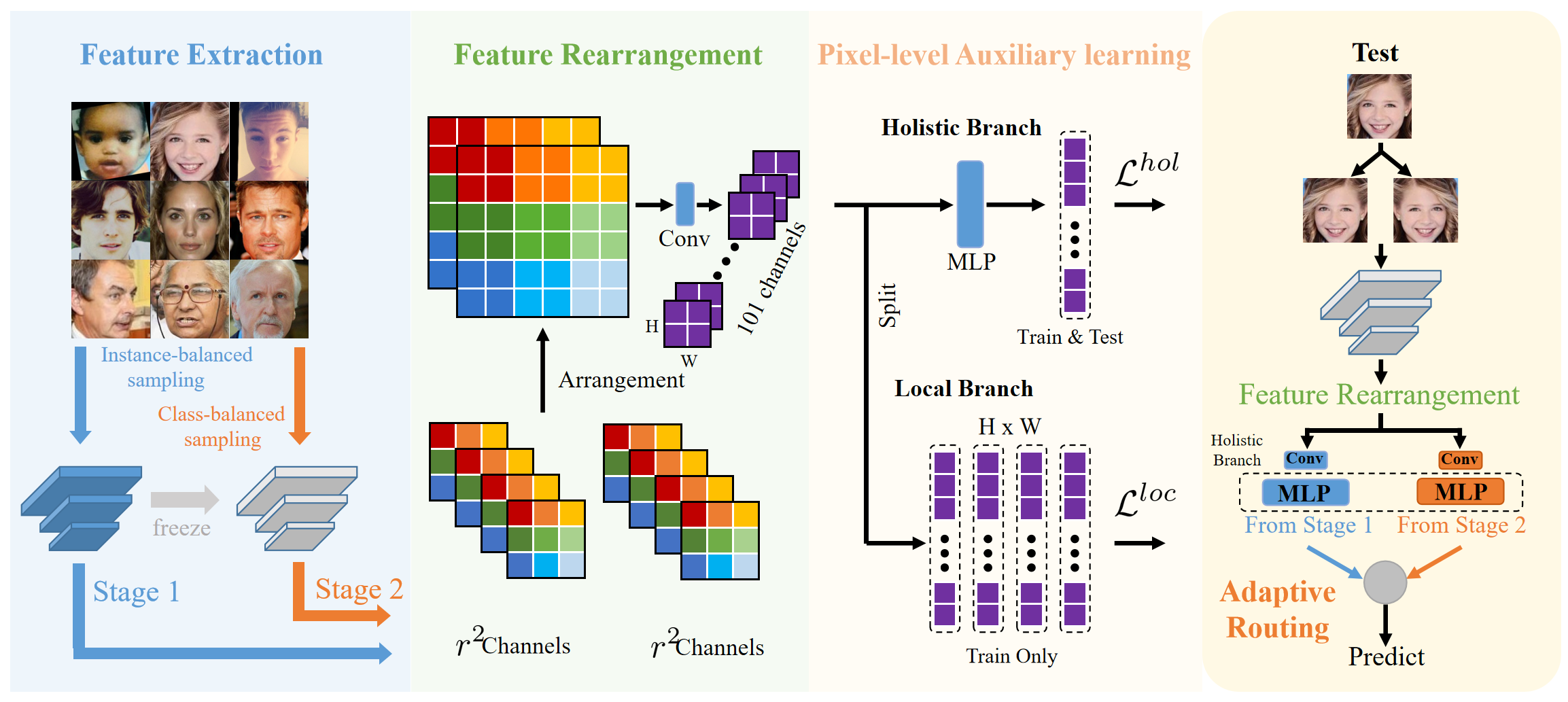}
\caption{The pipeline for the proposed GLAE. Given a face image, it will be fed to the feature extractor to generate a feature map. Then, the FR module will aggregate the dispersed information and the PA module will capture the local and holistic visual information to generate the predicted scores. In the Stage 1, the network is trained with instance-balanced sampling. In the Stage 2, the backbone of the first stage will be frozen and the rest of the network will be trained with class-balanced sampling. Finally, the appropriate classifier obtained from the two stages will be adaptively activated for predicting the input image.  
}
\label{fig_2}
\end{figure*}

\section{Proposed Work}
In this section, we first give out the traditional age estimation scheme in Sec. III-A. Then, we present the proposed Feature Rearrangement and Pixel-level Auxiliary learning components in Sec. III-B and Sec. III-C, respectively. Finally, we describe the proposed Adaptive Routing strategy in Sec. III-D. 

\subsection{Preliminaries and Motivation}
\label{sec3-a}
Following previous works~\cite{gao2018age,gao2017deep}, we train the network with a label distribution rather than a standalone label by considering the randomness in facial aging. Specifically, the corresponding label distribution will be set as a typical Gaussian distribution with a mean of the ground truth label.
Assume the sample is denoted as $({\bm x}, y)$ where ${\bm x}$ denotes the image and $y$ denotes the corresponding age label.
The $k$-th element of the label distribution is denoted as:
\begin{equation}\label{gau}
  {\bm z}^k = \frac{1}{\sqrt{2\pi}\sigma} exp{(-\frac{(k-y)^2}{2\sigma^2})}
\end{equation}
where $\sigma$ is the standard deviation and we set it as 1.
The symbol $k \in [0, ..., K]$ indicates an age index and $K$ is the maximum age.
We set $K=100$ and only consider the ages from 0 to 100 years.

Given an input image ${\bm x}$, we denote the prediction distribution produced by deep neural networks as ${\bm p}$,
where the $k$-th element ${p}^k$ represents the probability of classifying the input image to age $k$. In the training stage, the Kullback–Leibler (KL) divergence is adopted to minimize the distance between the ground truth label distribution ${\bm z}$ and the predicted distribution ${\bm p}$. The formula can be represented as:
\begin{equation}
    \ell_{kl}({\bm z}, {\bm p}) = \sum_{k=0}^{K}  {\bm z}^k  log \frac{{\bm z}^k}{{\bm p}^k} 
\label{L_sr}
\end{equation}

To obtain a specific age prediction, an expectation refinement is employed according to the work~\cite{rothe2015dex}. Specifically, the predicted age is denoted as: 
\begin{equation}
    \hat{y}=\sum_{k=0}^{K} k \cdot {p}^k
\end{equation}  

The expectation refinement takes the expectation of the output distribution as the final predicted age, which could enhance the stability and reliability of the prediction. Then, a $\ell_1$ regularization is further adopted to narrow the gap between the predicted age $\hat{y}$ and the ground truth label $y$,
which is defined as follows:
\begin{equation}
    \ell_{er}(y, \hat{y}) = | y - \hat{y}|  
\label{L_er1}
\end{equation}%
where $| \cdot |$ denotes $\ell_1$ distance.
To the end, the whole loss of training an age estimator with a label distribution and a $\ell_1$ regularization can be denoted as:
\begin{equation}\label{eq_teacher}
    \ell_{base} = \ell_{kl}({\bm z}, {\bm p}) +  \ell_{er}(y, \hat{y})
\end{equation}%
It is a simple yet effective method to train a model for age estimation,
which has been proven by many previous works~\cite{gao2018age,gao2017deep,bao2021lae}.
In this paper, we take it as the baseline model of our method (represented by $\ell_{base}$). 

Taking a closer look at these previous age estimators~\cite{gao2018age,pan2018mean,bao2021lae}, it is easy to find that the models usually consist of a feature extractor and a linear classifier, which are connected by a Global Average Pooling (GAP) layer. 
To be specific, the feature maps given by the last convolutional layer will be vectorized and fed into the classifier to generate predicted probabilities for age categories (101 categories in our case).
This extractor-classifier architecture is so widely used that few people would question whether the components in it are reasonable, especially the GAP. 
As the connection between the extractor and the classifier, a natural question is whether GAP completely retains the information carried by the feature maps and transforms it into the input form required by the classifier.
This consideration prompts us to rethink the GAP operational mechanism, 
which pool the last convolutional layer into a single vector by simply averaging. As a result, the pixel-level differences are erased, which inevitably results in the loss of the local and detailed features. Thus, the goal of our task is to design a new feature utilization mechanism to alleviate the loss of information. 
To this end, we propose the FR module to \textbf{aggregate the dispersed information} and the PA module to \textbf{fully capture the visual information} carried by the feature map. By combining FR and PA, we can learn high-quality representations for better General Age Estimation.

Apart from the aforementioned problems, the classifier in previous works~\cite{gao2018age,pan2018mean} usually shows poor performance in children and the elderly, which is mainly caused by the long-tailed distribution of age datasets. Specifically, the head classes (like adults) have more samples than the tail classes (like children and the elderly). The classifier trained on long-tailed datasets will inevitably show a prediction bias towards head classes. Previous works~\cite{bao2021lae} address the problem of long-tailed distribution by training a balanced classifier, which improves the performance of tail classes but also experiences a performance drop on head classes. Intuitively, only employing a single classifier can hardly perform well on both head classes and tail classes simultaneously. To this end, we propose an AR module to \textbf{select the appropriate classifier} for each input image. By selecting, we can avoid the performance trade-off, improving the performance of tail classes while maintaining the performance of head classes.

\subsection{Feature Rearrangement}

The core problem of aggregation is how to cluster the visual information carried by feature maps and how to properly transform the feature maps for subsequent classification. Moreover, considering the training efficiency, how to minimize the increase of parameters is also critical. Based on the above analysis, we propose the Feature Rearrangement module. 

Given a face image, it is fed into a conventional CNN (e.g., ResNet~\cite{he2016deep}) to produce the feature map ${\bm F} \in {\mathbb R}^{C\times H \times W}$ in the last convolutional layer,
where $C$, $H$ and $W$ indicate the channel number, height, and width of the feature map. Specifically, $C=512$, $H=7$ and $W=7$ if we take the classic ResNet-18 as the backbone with an input size of $224\times224$.
To aggregate the pixels corresponding to the same location, we propose a \textbf{re-arrangement} operation, whose detailed implementations are shown in Fig.~\ref{fig_2}.
The proposed operation re-arranges the adjacent $r^2$ channels to form a single channel, by collecting the pixels of the same location and placing them together in the re-arranged feature map.
This operation changes absolutle position of the pixels but maintains the sequence relationship. 
To be specific, the pixels of 1 to $r^2$ channels are placed from the upper left corner to the lower right corner in the corresponding re-arranged patch. 
Intuitively, the size of the re-arranged feature map is $C/r^2 \times Hr \times Wr$, where each pixel is spatially enlarged to $r \times r$ pixels, and the channels are reduced to $1 / r^2$.
We simply take ResNet18 as an example, where the output feature map ${\bm F} \in {\mathbb R}^{512\times 7 \times 7}$ 
will be re-arranged to ${{\bm F}}' \in {\mathbb R}^{2\times 112 \times 112}$ when $r = 16$. 

The re-arranged feature map ${{\bm F}}'$ has fewer channels but larger feature map size.
To construct the feature map with a proper size for subsequent classification, we design a special convolution layer to meet the demand. 
Specifically, the convolution layer is created based on the following considerations: 
(1) To fully capture the correlation of pixels of the same location in the input feature map ${\bm F}$, we set the convolution kernel size to $2r$ or $r$. Moreover, to capture the correlation of neighboring pixels, we set the stride to $r$ or $r/2$. In such a way, the information carried by the pixels corresponding to the same location and the correlation between neighboring pixels will be captured. 
(2) To generate corresponding channels for each age category for subsequent classification, the output channels of the convolution layer will be set to the number of categories. As a result, the fully connected layer can be omitted.

\subsection{Pixel-level Auxiliary learning}
After extracting the features of the final convolutional layer,
previous works~\cite{gao2018age,deng2021pml,bao2021lae} usually take a Global Average Pooling (GAP) layer to obtain a holistic feature vector, which is then fed into the classifier to produce the predicted results.
Although such a way is good at capturing global and holistic features,
it is defective in capturing meticulous and detailed information, which is also crucial in facial age estimation. In this sub-section, a Pixel-level Auxiliary learning (PA) is proposed to capture both holistic features and detailed information.
The proposed PA consists of two branches, 
where one generate predict vector from multiple hierarchy, and the other make predictions in a holistic way. We will introduce the details of the two branches in the following.

For the local branch, it first flattens the reconstructed features ${{\bm F}}' \in {\mathbb R}^{C'\times H' \times W'}$ into a series of vectors ${\bm f}_{h,w}^{loc} \in {\mathbb R}^{C'}$
w.r.t. $1 \leq h \leq H'$ and $1 \leq w \leq W'$, where each pixel corresponds to a vector. Then, a softmax layer is employed to obtain the predicted scores. 
For the pixel at position $(h,w)$, the predicted score for the age $k$ is denoted as:
\begin{equation}
    p_{h,w,k}^{loc} = \frac{ {\rm exp}({\bm f}_{h,w,k}^{loc})}{\sum_{k'}{\rm exp}({\bm f}_{h',w',k'}^{loc})}
\label{eq_softmax}
\end{equation}
The full predicted score is ${\bm p}_{h,w}^{loc} = [{ p}_{h,w,0}^{loc}, \cdots, { p}_{h,w,100}^{loc}]$,
and the predicted age $\hat y_{h,w}^{loc}$ can be inferenced by the expectation refinement introduced in Sec.III-A. Mathematically, the training loss employed for this branch can be denoted as:
\begin{equation} 
{\cal L}^{loc}
= \frac{1}{H'W'} \sum_{h=1}^{H'} \sum_{h=1}^{W'} (\ell_{kl}({\bm z}, {\bm p}_{h,w}^{loc} ) + \ell_{er}(y, \hat{y}_{h,w}^{loc}))
\label{eq_loss_lb}
\end{equation}

For the holistic branch, it first utilizes a MLP projection module to transformer the input feature map sized $C \times H\times W$ into a holistic feature embedding of $C$ dimensions. Given the reconstructed features ${{\bm F}}' \in {\mathbb R}^{C'\times H' \times W'}$, the MLP projection module could roughly denoted as:
\begin{equation} 
{\bm f}^{hol} = Proj( {{\bm F}}' )
\label{eq_proj}
\end{equation}
The employed MLP projection has two advantages. 
One is the feature reduction as mentioned above, 
which is a substitute for GAP but it avoids the loss of 
information caused by GAP. The other is to enhance features' capability and reduce the conflict between two employed branches (the conflict occurs if the classifiers of the two branches are constructed based on the same feature embedding).
After extracing the holistic features ${\bm f}^{hol}$, a softmax layer (similar to Eq.~\ref{eq_loss_lb}) is employed to produce the predicted scores ${\bm p}^{hol}$ and the corresponding predicted age $\hat{y}^{hol}$.
The loss for the holistic branch could be denoted as:
\begin{equation} 
{\cal L}^{hol}
= \ell_{kl}({\bm z}, {\bm p}^{hol} ) + \ell_{er}(y, \hat{y}^{hol})
\label{eq_loss_latter}
\end{equation}
Therefore, the summed loss of the two branches are represented as:
\begin{equation} 
{\cal L}^{sum}
= {\cal L}^{ loc} + {\cal L}^{ hol}
\label{eq_loss_sum}
\end{equation}
Note that the local classifiers in the local branch are only employed in the training stage, and all of them would be removed in the evaluation stage.

\subsection{Adaptive Routing}

To avoid the performance trade-off between head and tail classes, we propose to employ two classifiers for age prediction, where one is the vanilla classifier biased towards head classes and the other is the balanced classifier that performs well for tail classes. Considering the specialties of the two classifiers, the two classifiers are not activated concurrently for an input image. In other words, only the appropriate classifier is chosen for each input image to 
give full play to their advantages. To achieve this, we first decouple the training process into two stages as follows:

\begin{itemize}
\item \textbf{Stage 1:}
Train a strong and discriminative network using the proposed FR and PA. Specifically, the network is trained with an instance-balanced sampling way, which is the most common way to sample data. In instance-balanced sampling, all samples are treated equally and have the same probability of being sampled.
\item \textbf{Stage 2:}
Freeze the backbone and retrain a balanced classifier with a class-balanced sampling.
At this stage, all classes are treated equally and they have the same probability of being sampled. In this way, the retrained classifier is balanced and it would not be biased towards any class.
\end{itemize}

Two classifiers are obtained after the training process, namely a vanilla classifier and a balanced classifier. Then, an Adaptive Routing (AR) is proposed to select the appropriate classifier for each image, which could greatly improve the performance of tail classes while maintaining the performance of head classes. 
Given an image $\bm x$, both $\bm x$ and its fliped copy $\tilde{ \bm x }$ would be fed to the network and generate the predicted scores through the two classifiers.
For convenience, the correponding predicted scores for the vanilla classifier are denoted as ${\bm p}^{hol}$ and ${\tilde {\bm p}}^{hol}$,
and the scores for the balanced classifier are denoted as 
${\bm p}^{ban}$ and ${\tilde {\bm p}}^{ban}$. 
Then, we calculate the KL divergence between the predicted scores for each classifier, which is shown as follows:
\begin{equation}
  \Upsilon_1 =  \ell_{kl}( {\bm p}^{hol}, {\tilde {\bm p}}^{hol} ),\ 
  \Upsilon_2 =  \ell_{kl}( {\bm p}^{ban}, {\tilde {\bm p}}^{ban} )
\label{kl_two}
\end{equation}%

In the two-stage training, we use different sampling strategies. The classifier obtained in stage 1 performs well on the head classes (usually adults), and the classifier obtained in stage 2 performs well on the tail classes (usually children and elderly). Therefore, these two classifiers have better data augmentation robustness for samples in their respective class where they can output closer predictions. The higher the similarity between these two predictions, the better the model is able to recognize this input sample and has some resistance to interference. Conversely, a low similarity indicates that the model is less capable of recognizing that sample. We use the KL divergence as a measure of this similarity, that is, the smaller the KL divergence, the more similar. Thus, the classifier with the smaller KL divergence would be chosen for prediction. Specifically, the vanilla classifier is chosen if $\Upsilon_1 < \Upsilon_2$; otherwise, the balanced classifier is chosen.

\begin{algorithm}[t]   
\caption{Pseudo-code of our GLAE in a Pytorch-like style}   
\label{alg:Framwork}   
\begin{algorithmic}[1] 
\REQUIRE  ${\bm x} :$ samples, $f^b : $ backbone
\ENSURE $\hat{y} : $ final prediction
\STATE /* Stage 1 */
\REPEAT
\STATE Get samples with instance-balanced sampling.
\STATE /* Feature extraction. */
\STATE   ${{\bm F}}$ = $f^b$(${\bm x}$);
\STATE   /* Feature Arrangement. */
\STATE   ${{\bm F}}'$ = \emph{Conv}(\emph{re-arrangement}(${{\bm F}}$)) ; 
\STATE   /* Pixel-level Auxiliary learning. */
\STATE   ${\bm f}_{h,w}^{loc}$ = \emph{flatten}(${{\bm F}}'$) ;
\STATE   ${\bm f}^{hol}$ = \emph{Proj}(${{\bm F}}'$) ;
\STATE   /* Loss function. */
\STATE   Calculate ${\bm p}_{h,w}^{loc}$ as Equ.~\ref{eq_softmax} and ${\bm p}^{hol}$ as a similar way.
\STATE    Calculate ${\cal L}^{loc}$ as Equ.~\ref{eq_loss_lb} and  ${\cal L}^{hol}$ as Equ.~\ref{eq_loss_latter}.
\STATE   ${\cal L}^{sum}
= {\cal L}^{ loc} + {\cal L}^{ hol}$;
\STATE Optimize the network by ${\cal L}^{sum}$
\UNTIL the network converged.
\STATE /* Stage 2 */
\REPEAT 
\STATE   Get samples with class-balanced sampling.
\STATE   Freeze $f^b$ as backbone.
\STATE   Calculate ${\bm p}^{ban}$ as a similar way with ${\bm p}^{hol}$.
\STATE   Follow the procedure of Stage 1.
\UNTIL the network converged
\STATE /* Adaptive Routing */
\STATE Calculate $\Upsilon_1$ and $\Upsilon_2$ as Equ.~\ref{kl_two} 
\IF{$\Upsilon_1 < \Upsilon_2$}
\STATE ${\bm p} = {\bm p}^{hol}$
\ELSE
\STATE ${\bm p} = {\bm p}^{ban}$
\ENDIF
\RETURN $\hat{y}=\sum_{k=0}^{K} k \cdot {\bm p}^k$
\end{algorithmic}  
\end{algorithm}

\section{Datasets, Protocols and Metrics}
In this section, we introduce the details of the employed datasets, protocols, and evaluation metrics for both general and long-tailed age estimation.

\setlength{\tabcolsep}{8pt}
\begin{table}[t]
\caption{Statistics of Four Benchmark Datasets}
\centering
\resizebox{0.9\linewidth}{!}{
\begin{tabular}{c|cc|c}
\toprule[1pt]
Dataset           & Images  & Age range & SOTA  \\ \midrule[0.5pt]
Chalearn LAP 2015 & 7591    & 1-89      & 2.915~\cite{deng2021pml} \\
Morph II             & 55,134  & 16-77     & 1.74~\cite{zeng2020soft}  \\
CACD              & 163,446 & 16-62     & 4.35~\cite{tan2019deeply}  \\
MIVIA             & 575,073 & 1-81      &    --   \\ \bottomrule[1pt]
\end{tabular}
}
\label{tab:benchmark}
\end{table}

\subsection{Datasets and Protocols} 
Four datasets including Morph II~\cite{ricanek2006morph}, CACD~\cite{chen2014cross}, Chalearn LAP 2015~\cite{escalera2015chalearn} and MIVIA~\cite{greco2021effective} are used in our experiments and the corresponding statistical description are shown in Table~\ref{tab:benchmark}. Specifically, 
Morph II and CACD are collected from controlled and uncontrolled environments, respectively. Chalearn LAP 2015 and MIVIA are competition datasets for apparent age estimation. 

\textbf{Morph II}~\cite{ricanek2006morph} is one of the widely used public datasets, which contains 55,134 face images of 13,617 subjects, ranging from 16 to 77 years. We employ two protocols in our evaluation: (1) \textbf{Partial 80-20 protocol.} A subset of 5,493 face images of Caucasian descent following the work~\cite{tan_pami_2018} is selected. We randomly split the subnet into two parts: 80\% for training and 20\% for testing. (2) \textbf{Random 80-20 protocol.} As clarified in the works~\cite{gao2018age}, the whole dataset is randomly divided into two parts, with 80\% for training and 20\% for testing.

\textbf{CACD}~\cite{chen2014cross} 
is collected from the Internet Movie DataBase (IMDB) by searching celebrity name and year (2004–2013) as keywords. It contains more than 160 thousand images of 2,000 celebrities,
and the age labels range from 14 to 62.
The database contains much noise because the age was simply estimated by query year and birth year of that celebrity. We select 1,800 celebrities for training and 120 cleaned celebrities for testing, where the images are manually checked and the noise images are removed~\cite{tan_pami_2018}.

\textbf{Chalearn LAP 2015}~\cite{escalera2015chalearn} is a competition dataset for apparent age estimation (held in conjunction with ICCV 2015), and it collected 4,691 images. 
The annotations are labeled by at least 10 users and the average label is taken as the final annotation. Moreover, the dataset offers the standard deviation for each age label. 
We following the standard protocol~\cite{rothe2015dex,tan_pami_2018} to divide the dataset into training, validation, and testing subsets with 2,476, 1,136, and 1,079 images, respectively. We adopt the experimental settings of ~\cite{tan_pami_2018} for evaluation.

\textbf{MIVIA}~\cite{greco2021effective} is the competition dataset of \textit{CAIP Guess the Age Contest 2021}\footnote{https://gta2021.unisa.it/}. It contains 575K images extracted from the VGGFace2~\cite{cao2018vggface2} dataset. It is worth mentioning that the MIVIA Age Dataset is the largest publicly age dataset. 
Since the test set has not been released by the competition,
we randomly divide the MIVIA into a training set and a validation set at a ratio of 4:1 to validate the effectiveness of the proposed method.
Specifically, there are 458,752 images in the training set and 114,688 images in the validation set.

For general age estimation, the training and testing protocols are clarified as above.
For long-tailed age estimation, there are no corresponding protocols before.
In long-tailed visual recognition, one common way to evaluate the model is to construct a balanced test set, in which each category contains the same number of images.
However, due to the extreme long-tailed distribution of age datasets,
it is hard to partition a test set with each category containing adequate images (some classes only have 2 or 3 images). Therefore, we still follow the traditional protocols as mentioned above to split training and testing sets for long-tailed age estimation, but evaluate the model with a new balanced metric, which will be introduced in the following.

\subsection{Metrics}
For general age estimation, the Mean Absolute Error (MAE)~\cite{tan_pami_2018}
is usually adopted as the evaluation metric. It measures the mean absolute error between the predicted ages and the ground truth labels over the whole test set,
which is mathmatically denoted as:
\begin{equation}
MAE = \frac{1}{N}\sum_{i=1}^{N}  |y_i - \hat{y}_i|,
\label{eq_mae} 
\end{equation}%
where $N$ is the total number of images.

For long-tailed age estimation, a new metric called Class-wise Mean Absolute Error (CMAE) is proposed to consider all ages equally. To be specific, CMAE is represented as:
\begin{equation}\small
    CMAE = \frac{1}{M}\sum_{k=0,N_k > 0}^{K}( 
    \frac{1}{N_k}\sum_{i=1}^{N_k} 
    |y_{ik} - \hat{y}_{ik} | )
\label{eq_cmae}
\end{equation}%
where $N_k$ denotes the number of images of age $k$, 
and $M$ indicates the number of age categories containing images (\ie, $M = \sum \mathds{1}_{N_k > 0}$ where $\mathds{1}$ is the indicator function).

Moreover, following the works~\cite{tan2019deeply,bao2021lae}, we also employ  $\epsilon$-error and AAR~\cite{greco2021guess} to evaluate the models on Chalearn LAP 2015 and MIVIA datasets, respectively. The $ \epsilon$-error is defined as follows.
\begin{equation}
    \epsilon =  1 - \sum^N_{i=1}exp(-\frac{(\hat{y}_i-y_i)^2}{2\sigma_i^2})
\label{L_epsilon}
\end{equation}%
where $N$ is the number of samples, $ y_i$ and $\hat{y}_i$ are the age label and predicted age, $\sigma_i^2$ is the annotated standard deviation, respectively.
For AAR, its definition is as follows:
\begin{equation}\small
    AAR = max(0; 7-MAE) + max(0; 3-\sigma) 
\label{L_er}
\end{equation}%
\begin{equation}\small
    \sigma =  \sqrt{\frac{\sum^n_{j=1}(MAE^j - MAE)^2}{n}}
\label{L_er}
\end{equation}%
where MAE denotes the mean absolute error on the entire test set, 
$n$ denotes the number of age groups (10 years for a group), $MAE^j$ denotes the MAE that is computed over the samples whose real age is in $j$th age group. 

\section{Experiments}
In this part, we first present details on the experimental setup. Then, we thoroughly evaluate the impacts of each component in GLAE and compare our results with state-of-the-art methods on benchmark datasets in General Age Estimation and Long-tailed Age Estimation. Finally, we show the results of the long-tailed age estimation benchmark for three datasets.

\subsection{Implementation Details}
\subsubsection{Preprocessing} The images are aligned with five landmarks (including two eyes, nose tip, and two mouth corners) according to the work~\cite{zhang2016joint}. The faces are then cropped and resized to 224 × 224, and each pixel (ranged between [0,255]) is normalized by subtracting 127.5 and dividing by 128.  
For data augmentation, we use color jitter (a random sequence of brightness, contrast, saturation, hue adjustments) and RandAugment [3], where we set N=2 and M=9, where N denotes the number of transformations to apply, and M denotes the magnitude of the applied transformations.

\subsubsection{Training Details} All networks use \textbf{ResNet-18 (denoted by 11M) or ResNet-50 (denoted by 23M)} as the backbone and are pre-trained on ImageNet and optimized by SGD with Nesterov momentum. All models are implemented with Pytorch on 8 GTX 2080Ti GPUs. In the test stage, both the test image and its flipped copy are fed into the network, and the average prediction is used as the final prediction. The batch size is set to 32 for each GPU. As in the works~\cite{bao2021lae}, we use Onecycle~\cite{smith2019super} Scheduler as the optimizer scheduler.

\subsection{Ablation Study}
Here we conduct ablation experiments to verify the contributions of each module of the proposed GLAE on Random 80-20 protocol of the Morph II dataset.

 \setlength{\tabcolsep}{6pt}
\begin{table}[t]
\caption{Ablation study on Morph II datasets. Bold
indicates the best. $\downarrow$ ($\uparrow$) means the lower (higher) is better.}
\centering
\resizebox{0.6\linewidth}{!}{
\begin{tabular}{ccc|cc} 
\toprule[1pt]
\multicolumn{3}{c|}{ Method } & \multicolumn{2}{c}{ Morph II } \\ 
\midrule[0.5pt]
 $\ell_{base}$ & FR   & PA     & MAE$\downarrow$    & CMAE$\downarrow$  \\ \midrule[0.5pt]
&&&  1.74  &  2.30      \\
\checkmark&&& 1.51  & 1.86      \\ 
\checkmark&\checkmark&& 1.44   & 1.81    \\
\checkmark&\checkmark&\checkmark& \textbf{1.38}   &  \textbf{1.70}     \\ 
\bottomrule[1pt]
\end{tabular}
}
\label{ablation_module}
\end{table}

\subsubsection{Impact of Different Components}
As shown in Table~\ref{ablation_module}, we show the results of detailed ablation experiments. In these experiments, we use ResNet-18 as the backbone. The first row shows the result of using $\ell_{kl}$ as the loss function. As the modules are added, we present the model with $\ell_{base}$, the baseline model with FR module, and GLAE without AR, respectively. For $\ell_{base}$, it outperforms the model with $\ell_{kl}$ by 0.23 years in MAE and 0.44 years in CMAE. When adding the proposed modules to the network, the performance of MAE shows steady improvement. Moreover, with adding those modules one by one to the network, the CMAE also shows an upward trend. It shows that the model achieves a simultaneous improvement in General Age Estimation and Long-tailed Age Estimation. To make a fair comparison with other SOTA methods, the AR module will not be used in the General Age Estimation comparison. The validity of the AR module will be compared in Sec. V-D.

 \setlength{\tabcolsep}{8pt}
\begin{table}[t]
\caption{Ablation study of kernel size and stride. Bold
indicates the best. $\downarrow$ ($\uparrow$) means the lower (higher) is better.}
\centering
\resizebox{0.8\linewidth}{!}{
\begin{tabular}{c|cc|cl}
\toprule[1pt]
$r$                   & Kernel size & Stride & MAE & CMAE \\ \midrule[0.5pt]
\multirow{4}{*}{16} & 16     & 8      &  1.42   & 1.66     \\
                    & 16     & 16     &  1.40   &  1.74    \\
                    & 32     & 8      &  1.42   &  1.67    \\
                    & 32     & 16     &  1.40   &  1.80    \\ \bottomrule[1pt]
\end{tabular}
}
\label{ablation_r}
\end{table}

\subsubsection{Impact of Kernal Size and Stride}
In Sec. III-B, we introduce the proposed FR module, in which we use a convolution layer to construct the re-arranged feature map. To further analyze the effect of different kernel sizes and strides in the convolution layer, we conduct this ablation experiment in this subsection.
Since we use ResNet-18 as the backbone, the output feature map ${\bm F}_i \in {\mathbb R}^{512\times 7 \times 7}$ 
will be re-arranged to ${\tilde {\bm F}}_i \in {\mathbb R}^{2\times 112 \times 112}$ when $r = 16$. The reason we set $r$ to 16 is to maximize the aggregation of information scattered across different channels. In this situation, four combinations were evaluated, whose $(kernel\ size, stride)$ are $(r, r/2)$, $(r, r)$, $(2r, r/2)$, $(2r, r)$, as shown in Table~\ref{ablation_r}. First, the results of the four combinations differ less in MAE, and the main difference is concentrated in CMAE. Second, it is worth noting that when the stride is the same, the performance difference caused by different kernel sizes is mainly reflected in CMAE. When the kernel size is the same, the change of stride will cause a relatively large change in MAE and CMAE. When $(kernel\ size, stride)$ is set to $(16, 16)$, the model achieves the lowest MAE and lower CMAE compared to $(32, 16)$, so we use this combination in subsequent experiments. 

 \setlength{\tabcolsep}{6pt}
\begin{table}[t]
\caption{Ablation study of Projection layer depth. Bold
indicates the best. $\downarrow$ ($\uparrow$) means the lower (higher) is better.}
\centering
\resizebox{0.7\linewidth}{!}{
\begin{tabular}{c|c|cc}
\toprule[1pt]
Proj. depth & Param. & MAE  & CMAE \\ \midrule[0.5pt]
0           & 23M    & 1.14 &  1.27    \\
1           & 23.3M    & 1.12 & 1.20     \\
2           & 23.9M    & 1.10  & 1.14      \\ \bottomrule[1pt]
\end{tabular}
}
\label{ablation_mlp}
\end{table}

\subsubsection{Impact of MLP Depth}
In Sec. III-C, we introduce the proposed PA module, which has two branches: local branch and holistic branch. In the holistic branch, we basically follow the scheme of traditional age estimation but replace the GAP layer with MLP projection layers. The MLP projection layers transformer the input feature map sized $C \times H\times W$ into a holistic feature embedding of $C$ dimensions and avoids the loss of information caused by GAP. To further analyze the impact of the depth of projection layers on performance, we conduct this ablation experiment in this subsection. 

In these experiments, we use ResNet-50 as the backbone to evaluate the results of MAE and CMAE, as shown in Table~\ref{ablation_mlp}. First, we show the result when the depth is 0, which means only a linear layer is used to achieve the dimensionality reduction operations and no other activation function or batch normalization. It is not difficult to find that the introduction of MLP projection layers realizes the simultaneous improvement of MAE and CMAE. When the depth is 2, the model achieves an MAE of 1.10 years and a CMAE of 1.14 years. To make a fair comparison with other SOTA methods, we set the depth to 0 in subsequent experiments.

\subsection{General Age Estimation Comparison}
In this subsection, we compare the proposed method with the DEX~\cite{rothe2018deep}, AgeED~\cite{tan_pami_2018}, DRFs~\cite{shen2018deep}, DHAA~\cite{tan2019deeply}, BridgeNet~\cite{li2019bridgenet}, AVDL~\cite{wen2020adaptive}, POE~\cite{li2021learning}, PML~\cite{deng2021pml},ThinAgeNet~\cite{gao2018age}, CR-MTk~\cite{liu2020facial}, DOEL~\cite{xie2020deep}, AL-RoR-34~\cite{zhang2019fine} to verify the performance of GLAE on General Age Estimation.

\setlength{\tabcolsep}{4pt}
\begin{table}[t]
\caption{The MAE comparisons on Morph II dataset. Bold
indicates the best (* indicates the model is pretrained on external dataset and we use MS-Celeb-1M following the work~\cite{zeng2020soft}).}
\centering
\resizebox{0.9\linewidth}{!}{
\begin{tabular}{l|c|c|cc}
\toprule[1pt]
\multirow{2}{*}{Method} & \multirow{2}{*}{Year} & \multirow{2}{*}{Param.} & \multicolumn{2}{c}{Morph II} \\ 
\cmidrule{4-5} 
                    &    &  & Partial 80-20 & Random 80-20 \\ 
                        \midrule[0.5pt]
DEX~\cite{rothe2018deep}                & 2018   & 138M  & 3.25/2.68* & --       \\
AgeED~\cite{tan_pami_2018}               & 2018   & 138M & 2.93/2.52* & --       \\
DRFs~\cite{shen2018deep}                & 2018   & 138M & 2.91       & 2.17       \\
MV~\cite{pan2018mean}                   & 2018 & 138M & -- & 2.41/2.16* \\
DHAA~\cite{tan2019deeply}                & 2019   & 100M & 2.49       & 1.91      \\
AL-RoR-34~\cite{zhang2019fine}           & 2019 & 68M & 2.36 & --\\
BridgeNet~\cite{li2019bridgenet}           & 2020   & 138M & 2.38*      & 2.38*       \\ 
AVDL~\cite{wen2020adaptive}                & 2020   & 11M & 2.37*      & 1.94*       \\ 
CR-MTk~\cite{liu2020facial}              & 2020   & 60M & 2.31* & 2.30/2.15* \\
SR~\cite{zeng2020soft}                   & 2020  & 23M & -- & 1.74* \\
POE~\cite{li2021learning}                 & 2021   & 138M & 2.35*      &  --      \\ 
PML~\cite{deng2021pml}                 & 2021   & 21M & 2.31       & 2.15       \\ 
\midrule[0.5pt]
$\ell_{base}$      &  --  & 11M &      2.34*                    & 1.75*     \\
\textbf{Ours}                & -- & 11M &       2.34/2.29*                & 1.38* \\
\textbf{Ours}                & -- & 23M &      \textbf{2.00*}       & \textbf{1.14*}\\
\bottomrule[1pt]
\end{tabular}
}
\label{tab:morph}
\end{table}

\setlength{\tabcolsep}{12pt}
\begin{table}[t]
\caption{The MAE comparisons on CACD dataset. Bold
indicates the best (* indicates the model is pretrained on external dataset and we use MS-Celeb-1M following the work~\cite{zeng2020soft}).}
\centering
\resizebox{0.9\linewidth}{!}{
\begin{tabular}{l|c|c|c}
\toprule[1pt]
Method & Year & Param. & MAE$\downarrow$ \\ \midrule[0.5pt]
DEX~\cite{rothe2018deep}                & 2018   & 138M &4.79     \\
AgeED~\cite{tan_pami_2018}               & 2018   & 138M &4.68     \\
DRFs~\cite{shen2018deep}                & 2018   & 138M &4.63      \\
DHAA~\cite{tan2019deeply}                & 2019   & 100M &4.35     \\
CR-MTk~\cite{liu2020facial}              & 2020   & 60M &4.48   \\ \midrule[0.5pt]
$\ell_{base}$    &  --    & 11M    & 4.20*    \\
\textbf{Ours}   &  --      & 11M    &  4.34/4.14*   \\
\textbf{Ours}   &  --    & 23M    & \textbf{4.09*}    \\ \bottomrule[1pt]
\end{tabular}
}
\label{tab:cacd}
\end{table}

Table~\ref{tab:morph} shows the comparisons on the Morph II dataset. According to the results, our model achieves 2.34 years (without an external dataset) and 2.29 years (with an external dataset) under the Partial 80-20 protocol. Noticing that we only use a light network (\ie, ResNet-18) to achieve the best performance among all models except PML. Moreover, our GLAE achieves 2.00 years on a comparable model (\ie, ResNet-50), which outperforms all the previous state-of-the-art methods regardless of using the external dataset. Under Random 80-20 protocols, our GLAE achieves 1.38 years on ResNet-18 and 1.14 years on ResNet-50, which significantly outperforms the previous SOTA by 0.60 years. Such a performance increase is unprecedented, which demonstrates the effectiveness of the proposed method in a controlled environment. According to the results reported in Table~\ref{tab:cacd}, we compared our model with the state-of-the-art models on CACD. Our method GLAE achieves the lowest MAE of 4.14 years on ResNet-18 and 4.09 years on ResNet-50, which shows that our GLAE also works well in an uncontrolled environment.

As a competition dataset of apparent age estimation, the Chalearn LAP dataset is more special than other public datasets. We compare our model with the state-of-the-art method in Table~\ref{tab:clap}. Following the previous work~\cite{tan2019deeply}, we finetune the model on both training and validation sets after pretraining on a large additional age dataset, \ie, the IMDB-WIKI dataset or the MS-celeb-1M dataset. Moreover, we also report the performance on the validation set with only finetuning on the training set. Our GLAE achieves the lowest MAE of 2.852 years and the lowest $\epsilon$-error of 0.242 on the validation set, and achieves a decrease in the $\epsilon$-error by 0.015 on the test set, which is a large margin. The results on MAE and $\epsilon$-error both show that our proposed GLAE can also perform well on a small dataset. Another competition dataset is MIVIA, from Guess The Age Contest 2021, and has a large scale. For fair comparisons, we create the baseline model, which has the same architecture as us. Our method achieves the lowest MAE in each age group and achieves the highest AAR, which shows that our method still performs well on the large dataset. 



\setlength{\tabcolsep}{4pt}
\begin{table}[t]
\caption{The comparisons on the test set of Chalearn LAP 2015 dataset. Bold
indicates the best (* indicates the model is pretrained on external dataset and we use MS-Celeb-1M).
}
\centering
\resizebox{0.9\linewidth}{!}{
\begin{tabular}{l|c|c|cc|c}
\toprule[1pt]
\multirow{2}{*}{Method} & \multirow{2}{*}{Year} & \multirow{2}{*}{Param.} & \multicolumn{2}{c|}{Validation} &  Test \\ 
\cmidrule{4-6} 
              &   &     & MAE$\downarrow$    & $\epsilon$-error$\downarrow$   &  $\epsilon$-error$\downarrow$      \\ 
\midrule[0.5pt]
ARN~\cite{agustsson2017anchored}           & 2017 & 138M & 3.153* & -- & --        \\
DEX~\cite{rothe2018deep}           & 2018 & 138M & 3.252* & 0.282* & 0.265*          \\
AgeED~\cite{tan_pami_2018}         & 2018 & 138M & 3.21*  & 0.28* & 0.264*\\
ThinAgeNet~\cite{gao2018age}    & 2018 & 3.7M & 3.135* & 0.272* & --  \\
AL-RoR-34~\cite{zhang2019fine}     & 2019 & 68M  & 3.137* & 0.268* & 0.255* \\                          
DHAA~\cite{tan2019deeply}          & 2019 & 100M & 3.052* & 0.265* & 0.252*  \\
BridgeNet~\cite{li2019bridgenet}     & 2020 & 138M & 2.98*  & 0.26* & 0.255*\\
DOEL~\cite{xie2020deep}          & 2020 & 43M & 2.933* & 0.258* & 0.247* \\
PML~\cite{deng2021pml}           & 2021 & 21M & 2.915* & 0.243* & --\\
\midrule[0.5pt]
$\ell_{base}$  & -- & 11M   & 3.069*   &  0.265*   & 0.260*    \\
\textbf{Ours}  &  -- & 11M   & \textbf{2.852}*   &  \textbf{0.242}*   & \textbf{0.232}* \\ 
\bottomrule[1pt]
\end{tabular}
}
\label{tab:clap}
\end{table}

 \setlength{\tabcolsep}{6pt}
\begin{table}[t]
\caption{The comparisons on MIVIA dataset. Bold indicates the best.}
\centering
\resizebox{0.9\linewidth}{!}{
\begin{tabular}{c|ccc|cc}
\toprule[1pt]
\multirow{2}{*}{Method} & \multicolumn{3}{c|}{Group MAE} & \multicolumn{2}{c}{Overall} \\ \cmidrule{2-6} 
                        & 0$\sim$17 & 18$\sim$65  & 66$\sim$100 & MAE$\downarrow$  & AAR$\uparrow$   \\ \midrule[0.5pt]
 $\ell_{kl}$             & 3.84   & 1.81  & 2.64   &  1.86     &  6.64             \\
 $\ell_{base}$                      & 3.59  & 1.78 &2.50  & 1.83     & 6.92              \\
 \textbf{Ours}          & \textbf{3.41}   & \textbf{1.68}  & \textbf{2.31}  &\textbf{1.73}             &  \textbf{7.58}            \\ 
\bottomrule[1pt]
\end{tabular}
}
\label{tab:mivia}
\end{table}

A common problem associated with General Age Estimation evaluation is that the measured performance is often compromised by the long-tailed distributions prevalent in the dataset. As a metric that focuses on the average performance of all samples, MAE further exacerbates the proportion of head classes performance in it. However, under such evaluation metrics, it is impossible to measure the performance of age estimation algorithms across different age groups, especially in children and the elderly.

\setlength{\tabcolsep}{3pt}
\begin{table*}[t]
\caption{Long-tailed benchmark on benchmark datasets. All results are compared on CMAE. Bold indicates the best. Smaller $\Upsilon$ means that the predition distribution with a smaller KL divergence is chosen, and Bigger $\Upsilon$ means that the predition distribution with a bigger KL divergence is chosen.}
\centering
\resizebox{\linewidth}{!}{
\begin{tabular}{c|c|ccc|c|ccc|c|ccc|c|ccc|c}
\toprule[1pt]
\multirow{2}{*}{Method} & \multirow{2}{*}{Param.} & \multicolumn{4}{c|}{Morph II}  & \multicolumn{4}{c|}{CACD} & \multicolumn{4}{c|}{MIVIA} & \multicolumn{4}{c}{MIVIA-LT}   \\ \cmidrule{3-18} 
&   & 0$\sim$17  & 18$\sim$65  & 66$\sim$100 & ALL
& 0$\sim$17  & 18$\sim$65  & 66$\sim$100  & ALL
& 0$\sim$17  & 18$\sim$65  & 66$\sim$100  & ALL
& 0$\sim$17  & 18$\sim$65  & 66$\sim$100  & ALL \\ 
\midrule[0.5pt]
$\ell_{base}$  & 23M & 1.19 & 2.06 & 4.59  & 2.30 
&8.89 & 4.56 & -- & 4.96
& 4.76  & 1.80 & 2.81 & 2.62
& 4.93 & 1.73 & 2.66 & 2.58\\ 
\cmidrule{1-18} 
Vanilla  & 23M   & 0.85  & 1.39  & 1.55  & 1.39 
& 8.03  & 4.16 & -- & 4.48
& 4.15  & 1.70  & 2.53 & 2.38
& 4.16  & 1.66  & 2.36 & 2.32\\
Balanced  & 23M   & 0.97  & 1.45  & 0.69  & 1.34 
& 4.87  & 4.33  & --  & 4.37
& 2.85  & 1.99  & 1.98 & 2.17
& 2.92  & 1.97  & 2.02 & 2.18\\ 
\cmidrule{1-18} 
Bigger $\Upsilon$  & 23M   & 0.97  & 1.48  & 1.65  & 1.48 
& 7.10  & 4.27  & --  & 4.50
& 3.87  & 1.88  & 2.36 & 2.39
& 3.85  & 1.79  & 2.29 & 2.32\\
Smaller $\Upsilon$ (AR)  & 23M   & 0.85  & 1.40  & 0.59  & \textbf{1.27} 
& 4.90  & 4.18  & -- & \textbf{4.24}
& 3.08  & 1.73  & 2.10  & \textbf{2.09}
& 3.07  & 1.71  & 2.04  & \textbf{2.10}\\ 
\midrule[1pt]
\end{tabular}
}
\label{tab:mivia-lt}
\end{table*}

\subsection{Long-tailed Age Estimation Benchmark}
Considering the General Age Estimation evaluation, which cannot assess the balance of age estimation across age groups, we construct the first long-tailed age estimation benchmark to treat all classes equally and evaluate the age estimation performance across all ages. Since the Chalearn LAP 2015 dataset only contains 7,591 images, and the long-tailed phenomenon is relatively insignificant, we used three other age datasets for our experiments. For Morph II and CACD, which have some classes without samples or only two or three samples, it is not appropriate to force the construction of a balanced dataset and the new metric, CMAE, will be used for evaluation. For MIVIA, which has plenty of samples across all ages and has a notable long-tailed distribution, we apply both the balanced test set and CMAE for evaluation. 
Based on the actual situation of age estimation, we define children and teenagers (1$\sim$17), and the elderly (66$\sim$100) as tail classes, and adults (18$\sim$65) as head classes in our experiments. 
In addition, we use $\ell_{base}$ as the baseline for comparison. As seen in subsection C, \textit{\textbf{$\ell_{base}$ achieves comparable or even better performances than the previous SOTA methods on these three datasets}}, which can be used as a comparison to better indicate the effectiveness of the proposed method. Moreover, we compare the results of adopting different selection criteria in AR. Specifically, $small$ means that the classifier with a smaller KL divergence is chosen, and $big$ means that the classifier with a bigger KL divergence is chosen.

Table~\ref{tab:mivia-lt} shows the results of CMAE on three benchmark datasets. The first line shows the result of $\ell_{base}$ as a comparison, which can be simply regarded as a substitute for the existing state-of-the-art method. The second and third lines show the results of the vanilla and balanced classifiers. It is easy to find that the vanilla classifier performs better than the balanced classifier in head classes while the balanced classifier shows an advantage in the tail classes. The fourth and fifth lines show the results of adopting different selection criteria, it is obvious that our AR performs much better when we select the classifier with a smaller KL divergence, which shows the effectiveness of the selection strategy. Compared with the $\ell_{base}$, the proposed GLAE achieve an overall improvement in all age groups, which achieves the purpose of this paper. 
For the Morph II dataset, we use the Random 80-20 protocol for evaluation. It is worth noting that the CMAE performance on 0$\sim$17 is better than the CMAE performance on 18$\sim$65, which is caused by Morph II's relatively different data distribution. Specifically, the age range of the Morph II dataset is 16 to 77 years old, and the samples belonging to 16 and 17 years old are not less than the samples of adults. In other words, the tail classes of the Morph II dataset are mainly concentrated on the elderly. For CACD, the age range is 14 to 62 years old and the tail classes of the CACD dataset are concentrated on children. As the largest publicly age dataset, MIVIA has plenty of samples in each age category, but there is still a long-tailed distribution. The age range of the MIVIA is 1 to 81 years old, and the tail classes of the MIVIA dataset are concentrated on children and the elderly, which is shown in Fig.~\ref{fig_intro}. Apart from the similar test with other datasets, we also construct a balanced test set for evaluation, which is called MIVIA-LT. In this setting, each age class has 10 samples, and all of the samples belong to the original test set. The proposed GLAE also achieves excellent performance under MIVIA-LT and outperforms the $\ell_{base}$ by 0.48 years, which shows the proposed method can also perform well in a balanced test set.


\begin{figure*}[htb]
\centering
\includegraphics[width=7in]{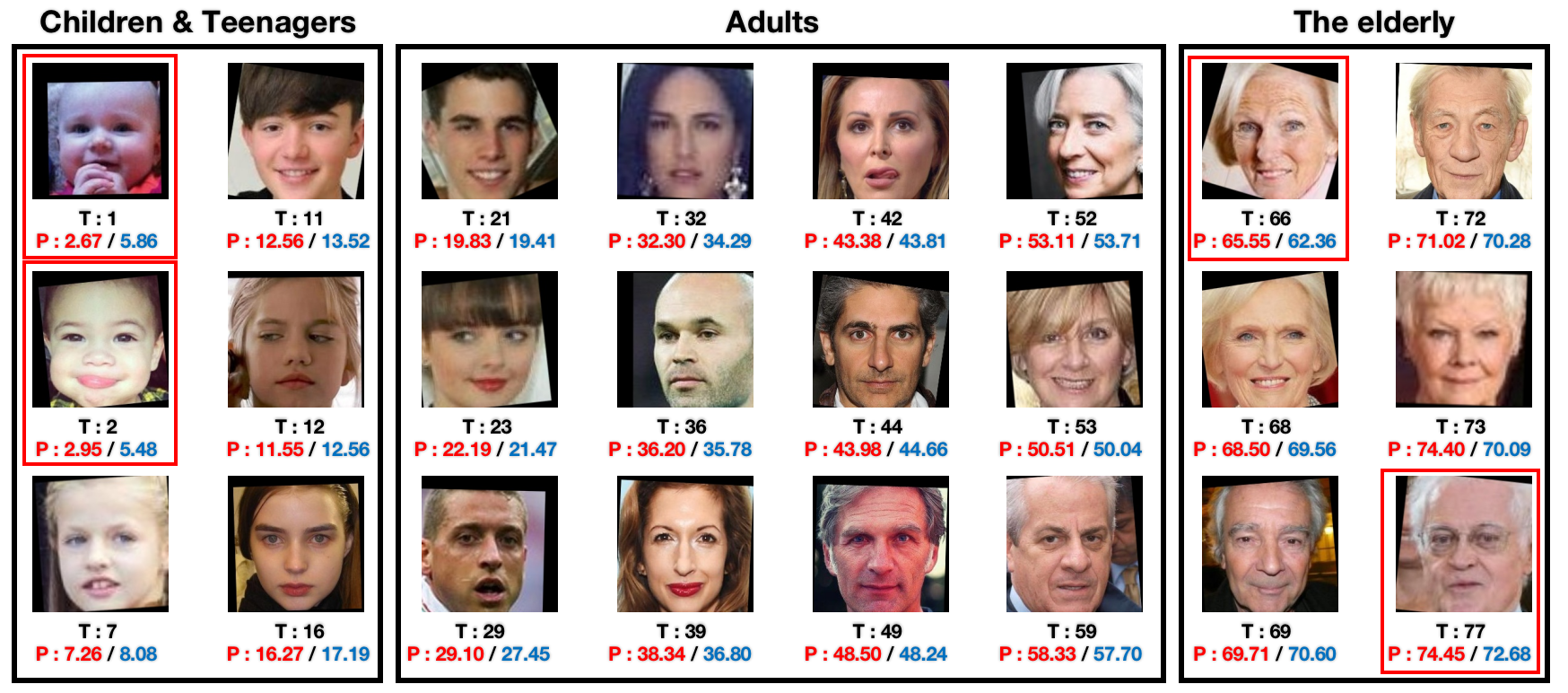}
\caption{Examples of age estimation results by the proposed approach on MIVIA dataset. The poor age estimation examples given by the baseline model are shown in red box (T: ground-truth age, P: predicted age, \textcolor{red}{red}: Ours, \textcolor{blue}{blue}: $\ell_{base}$).}
\label{fig_image}
\end{figure*}

\section{Visualization and Discussion}

\subsection{Qualitative Results} 
To better demonstrate the effectiveness of our GLAE intuitively, we show examples of age estimation results on the MIVIA dataset. We randomly select several images of different ages from MIVIA's test set and arranged them in ascending age order, and the results are shown in Fig~\ref{fig_image}. The predicted results of $\ell_{base}$ (blue), our GLAE (red) and the ground truth labels (black) are shown below the image. As you can see, our GLAE shows excellent performance in all age groups, whether children, adults, or the elderly. Specifically, compared to the $\ell_{base}$, which has been proved to be superior or similar to previous SOTA methods in Sec. V-C, our method shows steady improvement in adults. As for children and the elderly, the poor predictions ($>$ 3 years) of the $\ell_{base}$ are shown in the red box, which indicates that the $\ell_{base}$ faces a large performance decline in children and the elderly. It is obvious that our GLAE still performs well in these samples, and achieves better performance than the $\ell_{base}$, which indicates the effectiveness of our GLAE.

\begin{figure}[htb]
\centering
\includegraphics[width=3.5in]{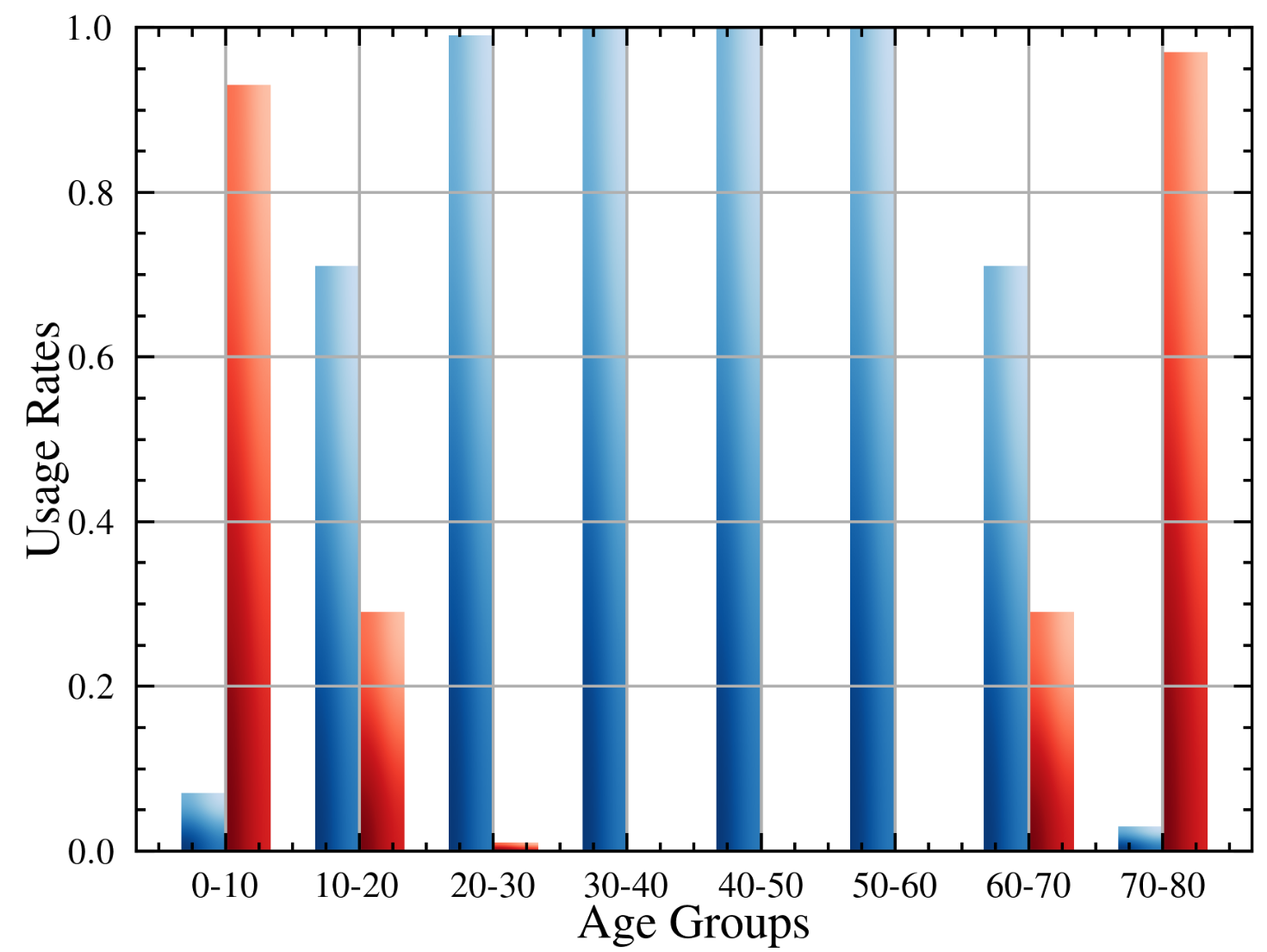}
\caption{The usage ratios of Vanilla (Blue) and Balanced (Red) classifiers in different age group.}
\label{fig_ar}
\end{figure}

\subsection{Details of Adaptive Routing}
To better illustrate the effectiveness of the AR module, we counted the number of times the Vanilla and Balanced classifiers were adaptively activated in each age group. Due to the excessive difference in sample size between the different age groups, we used the usage rates to represent the use of the Vanilla and Balanced classifiers. The usage ratios are tabulated in Fig. 4. It can be seen that when the input samples are adults, the vast majority of them activate the Vanilla classifier for the final prediction. When the input samples are children and the elderly, the number of activations of the Vanilla classifier decreases significantly, and the Balanced classifier starts to dominate in making the final prediction. This is in line with our expectations and further demonstrates that our adaptive routing module indeed selects the appropriate classifier for prediction.

\subsection{Generalization of FR and PA}
The role of the feature rearrangement module and pixel-level auxiliary learning module is to improve feature quality by reducing information loss caused by the GAP, thereby improving classification accuracy. To better demonstrate the effectiveness of these two modules, we conducted experiments on a general classification task. It should be noted that we strictly followed the training example provided by the PyTorch official website (https://github.com/pytorch/examples/tree/main/imagenet). We trained the original ResNet50 model from scratch for 90 epochs without using any additional training strategies. Therefore, the baseline accuracy is not high. Our model, on the other hand, only added the two modules we designed without any other changes and did not increase the number of parameters, making it a fair comparison. The results are shown in Table~\ref{tab:imagenet}. It can be seen that our method outperforms the original model, further verifying the effectiveness of the proposed method.

\setlength{\tabcolsep}{12pt}
\begin{table}[t]
\caption{The comparisons on Imagenet. Bold indicates the best.}
\centering
\resizebox{0.8\linewidth}{!}{
\begin{tabular}{c|cc}
\toprule[1pt]
method   & \#Param. & ImageNet top-1 acc. \\ \midrule[0.5pt]
ResNet50 & 23M      & 75.8                \\
Ours     & 23M      & \textbf{76.6}                \\ \bottomrule[1pt]
\end{tabular}
}
\label{tab:imagenet}
\end{table}

\subsection{Future Work}
In this paper, we propose FR and PA modules based on the existing extractor-classifier architecture, with the goal of addressing the information loss caused by GAP and effectively improving the performance of facial age estimation. We believe that these two modules are not only applicable to facial age estimation but also can be plug-and-play in other similar classification tasks. Thus, in the future, we will explore the effectiveness of the proposed modules on other classification tasks.

In addition, although the introduction of the AR module effectively improves the performance of the model on the tail classes, the role of the AR module is more in the adaptive selection, and the performance on the tail classes depends more on the Balanced classifier. Therefore, how to get a better Balanced classifier, or rather, a classifier with better performance in the tail class, is a direction worthy of our further exploration in the future.


\section{Conclusion}
In this paper, we propose a unified framework, named GLAE (General vs. Long-Tailed Age Estimation), for facial age estimation, which performs well on both General Age Estimation and Long-tailed Age Estimation. Our method is two-fold. The Feature Rearrangement and Pixel-level Auxiliary learning module provide a better feature utilization mechanism for fully capture the visual information carried by the feature map. By combining FR and PA, we can learn high-quality representations for better general age estimation. The Adaptive Routing module provides an adaptive strategy for selecting the appropriate classifier for each input image. By selecting, we can avoid the performance trade-off between head and tail classes. Moreover, we propose a new evaluation criterion named CMAE and several protocols to evaluate the performance for long-tailed facial age estimation. Extensive experiments on multiple age benchmark datasets, including CACD, Morph, MIVIA, and Chalearn LAP 2015, indicate that the proposed method outperforms the state-of-the-art approaches significantly.

\bibliographystyle{IEEEtran}
\bibliography{egbib}

\vfill

\end{document}